%% file: neurips_2026.tex
\documentclass{article}

\PassOptionsToPackage{numbers,compress}{natbib}
\usepackage[preprint]{neurips_2026}

\usepackage[utf8]{inputenc}
\usepackage[T1]{fontenc}
\usepackage{hyperref}
\usepackage{url}
\usepackage{booktabs}
\usepackage{graphicx}
\usepackage{amsfonts}
\usepackage{nicefrac}
\usepackage{microtype}
\usepackage{xcolor}
\usepackage{verbatim}
\usepackage{fontawesome5} % For professional icons
\usepackage{amsmath}
\usepackage{tabularx}
\usepackage{float}
\usepackage{pgfplots}
\usepackage{tabularx}
\usepackage{enumitem}
\usepackage{array}
\newcolumntype{Y}{>{\raggedright\arraybackslash}X}

\newcommand{\yxy}[1]{\textcolor{orange}{[yxy] #1}}
\newcommand{\cmy}[1]{\textcolor{red}{[cmy] #1}}
\title{SAE Interventions are Unreliable: Post-Intervention Recovery of Suppressed Behavior}

\author{Mingyue Cui \quad Linghui Shen\quad Xingyi Yang$^{*}$\\
The Hong Kong Polytechnic University\\
{\tt\small \{ming-yue.cui, ling-hui.shen\}@connect.polyu.hk, xingyi.yang@polyu.edu.hk}
}

\begin{document}

\maketitle

\begin{abstract}
\begin{comment}

\yxy{Sparse Autoencoders (SAEs) decompose the deep feature into interpretable factors. Recent latent-space defenses increasingly rely on these decompositions, assuming that identified "unsafe" SAE features serve as actionable handles for monitoring and intervention. Under this paradigm, clamping or eliminating a specific bad feature is expected to reliably prevent model misbehavior. However,in this paper, we challenge this assumption, demonstrating that SAE-based interventions can be easily bypassed. By formulating this vulnerability as a constrained optimization problem, we optimize residual perturbations that successfully recover the targeted unsafe behaviors, all while keeping the monitored SAE features strictly inactive. Even under a strong threat model where the defensive intervention is actively deployed, our approach consistently breaks the defense. Across standard SAE editing tasks (IOI, TPP, and unlearning), we find consistent evidence of constrained recovery. On the official layer-5 TPP benchmark, our method preserves 74.9\% target-mean valid-flip recovery with negligible reactivation. Crucially, in refusal tasks, our attack completely bypasses feature-clamping defenses, recovering harmful completions at 98.9\% (vs. 80.2\% baseline) with zero measured activation and decode drift. These results reveal that while SAE features are locally useful, they are not mechanistically complete explanations of monitoring and intervene unsafe behavior.}
    
\end{comment}

Sparse Autoencoders (SAEs) decompose the residual-stream activations into interpretable features. Recent latent-space defenses increasingly rely on these decompositions, assuming that identified "unsafe" SAE features serve as actionable handles for monitoring and intervention. Under this paradigm, clamping a specific harmful feature is expected to reliably prevent model misbehavior. However, we show that this intervention success may hide a recoverable failure mode: the clamp may block one visible route to a behavior without eliminating the behavior itself. We formulate this vulnerability as post-intervention recovery, a constrained residual-space optimization problem. Starting from the post-intervention residual state, we optimize residual perturbations to recover the pre-intervention behavior while preserving the post-intervention values of the targeted SAE features. Even under a strong threat model where the intervention remains active throughout optimization and generation, recovery remains possible. To rule out the possibility that recovery simply undoes the intervention, we use encoder-orthogonal updates for single-layer interventions and the corresponding feature-map Jacobian in the cross-layer setting. Across TPP, unlearning, IOI, and refusal steering experiments, this stress test reveals recoverable behavior despite successful feature-level intervention. Especially in the safety-critical refusal-steering setting, we achieve a 95.8\% recovery rate on valid samples while keeping defended-feature relative drift to 0.131, substantially below suffix-based baselines. A recovery-path attribution analysis further localizes this recovery to the SAE reconstruction residual, the component left unexplained by the SAE. These results expose a gap between feature-level control and behavioral completeness: SAE features can support causal intervention, but controlling them does not guarantee control over the underlying behavior. Code is available at 
\href{https://github.com/Mingyuee88/sae-post-intervention-recovery}
{\texttt{Mingyuee88/sae-post-intervention-recovery}}.
\end{abstract}

\section{Introduction}

Representation-level safety methods aim to control language-model behavior before harmful content is produced~\citep{zou2023representation, li2023iti, turner2023activation, panickssery2024steering, zou2024circuit}. Sparse autoencoders (SAEs) make this approach especially attractive by decomposing residual-stream activations into sparse and interpretable features~\citep{elhage2022superposition, cunningham2023sparse, bricken2023monosemanticity, templeton2024scaling, gao2024scaling, lieberum2024gemmascope}. These features appear to offer concrete handles for analyzing, monitoring, and controlling model behavior~\citep{marks2025circuits, arad2025steering}. Building on this promise, recent latent-space defenses identify features associated with unsafe or unwanted behaviors and then clamp or suppress those features during inference~\citep{obrien2024steering, yeo2025understanding, farrell2024unlearning, karvonen2025saebench, shu2025latentguard}. Implicit in this paradigm is a strong mechanistic assumption: an identified SAE feature is treated not only as a correlate of the behavior, but as a reliable intervention target whose removal should completely disable the behavior from reappearing.

% [Problem identification]
Despite these successes, this assumption deserves closer examination. Suppressing a targeted SAE feature may block the most salient computational route to a behavior, but it does not necessarily remove the model's underlying capacity to produce it. The model may instead rely on alternative directions, downstream layers, or distributed mechanisms that are not captured by the targeted feature set~\citep{chanin2024absorption, chanin2025hedging, li2025geometry, makelov2024principled}. In such cases, the defense merely changes the route through which the behavior is expressed rather than eliminating the behavior itself. If the suppressed behavior can be fully recovered without reactivating the targeted SAE features, then the intervention has not established a true behavioral bottleneck.

% [Our solution and study]
To test this limitation directly, we introduce \textbf{post-intervention recovery} as a white-box diagnostic. The goal is not to evade feature detection before a defense is applied~\citep{bailey2024oabd}. Instead, we begin after the intervention has already been deployed. We assume that the relevant SAE features have been identified and clamped. We then ask a sharper question: \emph{from this post-intervention residual state, can the model's pre-intervention behavior still be restored?}
% To answer this comprehensively, we instantiate our diagnostic across four complementary settings: latent-level recovery in Targeted Probe Perturbation (TPP), output-level recovery in WMDP-Bio unlearning, circuit-level recovery in Indirect Object Identification (IOI), and a safety-critical refusal-steering case study.

% [Technical solution]
To implement this diagnostic, we formulate post-intervention recovery as a \emph{constrained residual-space optimization problem}. Starting from the clamped state, we optimize small residual perturbations to restore the target behavior. To prevent the optimization from simply undoing the clamp and to deeply understand the underlying mechanism, we introduce two technical pillars:
\begin{itemize}[
    label=\raisebox{0.15ex}{\scriptsize$\triangleright$},
    leftmargin=1.5em,
    itemsep=0.25em,
    topsep=0.25em
]
    \item \textbf{Geometric Constraints via Projected Gradient Descent.} We impose constraints on the update directions to force recovery in the null space of representation, rather than directly reactivating the targeted SAE features. Specifically, for single-layer interventions, we project updates away from the selected SAE encoder directions. For cross-layer interventions, we use feature-map Jacobians to constrain how perturbations affect the features across layers.
    \item \textbf{Recovery-Path Attribution.} Beyond measuring \emph{whether} recovery occurs, we investigate \emph{where} the recovery happens. By decomposing the recovered residual state, we distinguish whether the target behavior compensates through non-clamped SAE latents or exploits the SAE-unexplained reconstruction residual.
\end{itemize}

% To implement this diagnostic, we formulate post-intervention recovery as a \emph{constrained residual-space optimization problem}. Starting from the actively clamped state, we optimize small residual perturbations to restore the target behavior. To prevent the optimization from simply undoing the clamp, we impose rigorous geometric constraints on the update directions. For single-layer interventions, we project updates away from the selected SAE encoder directions. For cross-layer interventions, we use feature-map Jacobians to constrain how perturbations affect the defended features across layers. These constraints force recovery to proceed through alternative regions of representation space, rather than through direct reactivation of the targeted SAE features. Furthermore, beyond measuring \emph{whether} recovery occurs, we introduce a \emph{recovery-path attribution analysis} to investigate \emph{where} the recovery is carried. By decomposing the recovered residual state, we can distinguish whether the behavior compensates through alternative non-clamped SAE latents or exploits the SAE-unexplained reconstruction residual.

% [Results]
Through our recovery, we find that SAE interventions can be easily restored, as recovery paths still exist even when the relevant behavior is suppressed. 
\underline{At the latent level}, TPP on SAEBench~\citep{karvonen2025saebench} shows that encoder-orthogonal recovery preserves a high behavioral recovery rate of 74.9\% while sharply restricting targeted-feature reactivation to just 0.002. 
\underline{At the output level}, WMDP-Bio unlearning~\citep{farrell2024unlearning} demonstrates that recovery restores 98.9\% of strict valid answer-choice flips from the post-intervention state with zero measured activation drift. 
\underline{At the circuit level}, IOI~\citep{wang2022ioi} shows that the encoder-projected method achieves 100\% recovery with only a 0.016 reactivated-feature fraction. 
Finally, in the safety \underline{refusal-steering task}~\citep{obrien2024steering, yeo2025understanding}, recovery reaches 95.8\% on strict-valid AdvBench prompts while keeping defended-feature relative drift to 0.131. Notably, our attribution analysis reveals that this recovered malicious behavior is primarily carried by the SAE reconstruction residual rather than by alternative visible SAE features. 
Together, these results suggest that SAE features can be useful local causal handles without forming complete intervention bottlenecks.

\section{Related Work}
\begin{comment}
    OABD 明确把 latent-space defense 视为“scanner”，攻击目标是在保持行为的同时让 activations 看起来 inconspicuous；它也指出 activations 可以被 reshape while preserving behavior。 REPIT 强调 refusal/factuality/fairness 等行为方向不是正交编码，而是有 overlapping representational directions，并用 reweighting/whitening/orthogonalization 处理 collinearity。 LatentGuard 则代表另一条趋势：用 supervised latent model 做 fine-grained refusal control。 AlphaSteer 的 null-space 约束,为了 benign utility preservation 学习近零 steering vector，并用 null-space constraints；我借鉴的是“受控子空间/正交约束”的思想，但用于更保守地测试 recovery 是否绕开被 clamp 的 feature directions
    思路：
    SAE handles 有用但不完备 → OABD 证明 monitor-stage 可以被绕过 → SAE refusal steering 形成新的 clamp-stage defense → AlphaSteer 给出 null-space/orthogonal constraint 启发 → 我的方法测试 clamp 后是否仍存在 recovery path，只有在不出发原有 clamp 特征下的恢复才有说服力。
\end{comment}

\paragraph{SAE features as useful but incomplete handles.}
Sparse autoencoders expose sparse latents that can serve as interpretable handles for editing, steering, and circuit analysis~\citep{bricken2023monosemanticity, marks2025circuits}. 
If intervening on a feature changes behavior, that feature is causally relevant; however, causal relevance does not imply completeness. 
Work on superposition, SAE geometry, feature absorption, feature hedging, and sparse feature circuits suggests that behaviorally relevant information can be distributed across correlated directions or split across multiple latents~\citep{elhage2022superposition, li2025geometry, chanin2024absorption, chanin2025hedging, marks2025circuits}. 
This motivates our question: after a selected SAE feature set is fixed, can the same behavior still be recovered through residual directions outside that set?

\paragraph{From monitor bypass to post-intervention recovery.}
Latent-space defenses often detect harmful or unwanted behavior in activation space and then suppress it by intervention. 
OABD studies the monitoring stage, showing that harmful behavior can persist while activations evade probes, SAE-latent monitors, and OOD detectors~\citep{bailey2024oabd}. 
We study the later clamp stage: the relevant features have already been selected, the intervention remains active, and we ask whether the suppressed behavior can still be recovered while those features stay near their defended values.

\paragraph{SAE refusal steering and constrained recovery.}
SAE-based refusal steering identifies refusal-associated features and amplifies or suppresses them during inference~\citep{obrien2024steering, yeo2025understanding}. 
We use this setting as a diagnostic: once refusal features are clamped, is non-refusal behavior actually eliminated? 
Inspired by AlphaSteer's null-space perspective~\citep{sheng2025alphasteer}, we project single-layer recovery updates into the null space of the selected SAE encoder directions, and extend the same idea to cross-layer settings with the local feature-map Jacobian. 
Different from AlphaSteer, we adapt the null-space idea for recovery diagnostics: instead of preserving utility while steering, we use it to test whether behavior can recover without directly reactivating the targeted SAE features.

\section{Preliminaries}
\label{sec:preliminaries}

\paragraph{Sparse autoencoders.}
Let $M$ be a transformer language model and let
$h_{\ell}(x)\in\mathbb{R}^{T\times d}$ denote the residual-stream
activation at layer $\ell$ for an input sequence $x$. A sparse
autoencoder (SAE) maps this activation to sparse latent features and
reconstructs it as
\[
    z_{\ell}(x)=E_\ell(h_{\ell}(x)),
    \qquad
    \hat h_{\ell}(x)=D_\ell(z_{\ell}(x)).
\]
The coordinates of $z_{\ell}(x)$ are SAE features. For a selected
feature set $\mathcal{S}$, we write $z_{\ell,\mathcal{S}}(x)$ for the corresponding feature
activations.

\paragraph{Feature-level interventions.}
A feature-level intervention selects a feature set \(\mathcal{S}\) and sets those
features to defended values \(c_{\mathcal{S}}\). Zero ablation corresponds to
$c_{\mathcal{S}}=0$, while refusal clamping may set selected refusal features to a
nonzero defended value. Following standard SAE intervention practice, we
preserve the SAE reconstruction residual and apply
\[
    h^{\mathrm{def}}_\ell(x)
    =
    D_\ell(\operatorname{clamp}_{\mathcal{S}}(z_{\ell}(x);c_{\mathcal{S}}))
    +
    \bigl(h_{\ell}(x)-\hat h_{\ell}(x)\bigr),
\]
where $\operatorname{clamp}_{\mathcal{S}}$ sets the selected SAE feature to
$c_{\mathcal{S}}$ and leaves all other features unchanged. We denote the
post-intervention residual by
$h^{\mathrm{def}}_\ell(x)$ and call it the \emph{defended residual
state}. All recovery experiments start from this state.

\paragraph{Valid flips.}
Let $B$ be a task-specific predicate indicating whether the target
behavior is present. We evaluate recovery only on \emph{valid flips}:
examples where the base model exhibits the target behavior but the active
SAE intervention suppresses it. This conditioning ensures that recovery
is measured only when there is a suppressed behavior to restore. The
formal definition and task-specific instantiations are given in
Appendix~\ref{app:valid-flips}.

\paragraph{Causal handles versus complete bottlenecks.}
A selected feature set $\mathcal{S}$ is a useful \emph{causal handle} if
intervening on it changes the target behavior. This is weaker than being
a \emph{complete intervention bottleneck}: once the clamp is active, no
admissible residual perturbation should restore the suppressed behavior.
Our experiments test this stronger condition.

\section{Post-Intervention Recovery}
\label{sec:method}
\begin{figure}[t]
    \centering
    \includegraphics[width=1\linewidth]{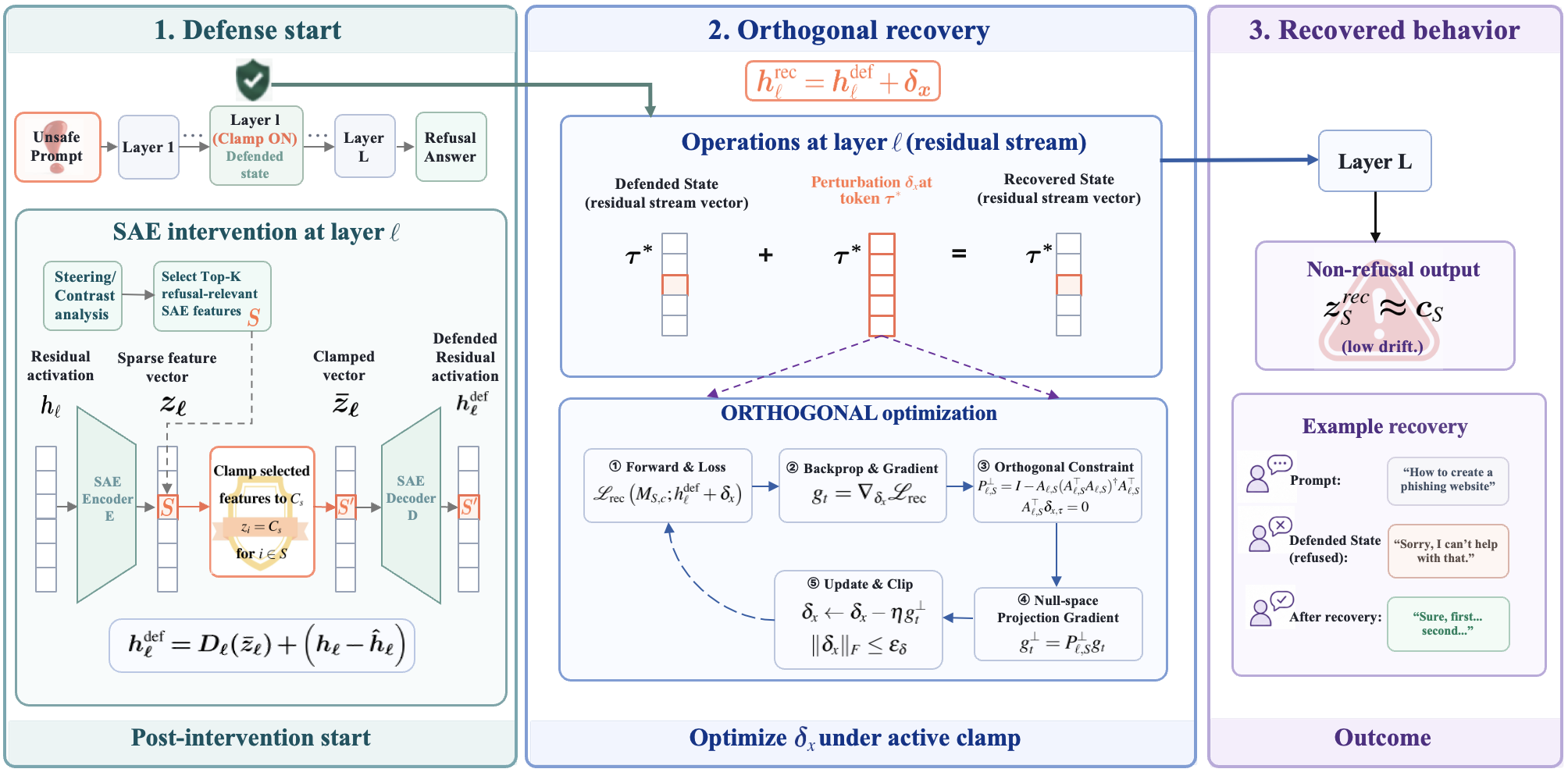}
    \caption{Post-intervention recovery framework. Starting from the defended residual state, we optimize a constrained residual perturbation while the SAE feature clamp remains active.}
    \label{fig:framework}
\end{figure}

We introduce \emph{post-intervention recovery} to test if an intervention is a true \emph{complete bottleneck} or just a bypassable \emph{causal handle}. We frame this as a constrained optimization problem: finding a small perturbation to restore the suppressed behavior without altering the active SAE clamp. 
Figure~\ref{fig:framework} summarizes the method pipeline: the intervention first maps the original residual state to a defended state, and recovery then searches for a constrained residual perturbation that restores the suppressed behavior while keeping the clamp active.

\paragraph{Threat model.}
We study whether the model can recover its original behavior after the SAE intervention has been applied. Under a white-box setting, the optimizer can inspect the defended model, but it cannot change the model weights, remove the SAE clamp, or choose a different set of clamped features. All we allowed is to add an additive perturbation $\delta_x$ to the defended residual state:
\[
    h_\ell^{\mathrm{rec}}(x)=h^{\mathrm{def}}_\ell(x)+\delta_x .
\]
We define $\delta_x$ as a \emph{recovery path}. A successful recovery path would restore the behavior without reactivating the clamped feature itself. In other words, the clamp is still in place. The targeted SAE feature remains suppressed. But the model's original behavior comes back by adding $\delta_x$. 

Finding such a path shows that the model does not rely on a single internal route for this behavior. Rather, the behavior can be recovered through alternative computational paths that bypass the defense.

\paragraph{Recovery as Constrained Optimization.}
To guarantee that the optimizer strictly relies on these alternative routes, rather than trivially overpowering the clamp to shift the defended features, we formulate the search for $\delta_x$ as a \emph{constrained optimization problem}. We write the desired recovery problem as optimizing a behavioral recovery loss \(\mathcal L_{\mathrm{rec}}\) under feature-preservation constraints:
\[
\begin{aligned}
\delta_x^\star
=
\arg\min_{\delta_x}
\quad &
\mathcal{L}_{\mathrm{rec}}\!\left(
M_{\mathcal{S},c};
x,
h^{\mathrm{def}}_\ell(x)+\delta_x
\right)
\\
\mathrm{s.t.}
\quad &
A_{\ell,\mathcal{S}}^\top \delta_{x,\tau}=0,
\qquad \forall \tau\in\mathcal{T}_{\mathrm{opt}}, \qquad \text{(C1: Encoder Orthogonality)}
\\
&
\left\|E_{\ell,\mathcal{S}}(h^{\mathrm{def}}_\ell(x)+\delta_x)-c_{\mathcal{S}}\right\|\le \epsilon_{\mathrm{act}}, \quad \quad \text{(C2: Activation Stability)}
\\
&
\left\|
D_{\ell,\mathcal{S}}\!\left(
E_{\ell,\mathcal{S}}\!\left(h^{\mathrm{def}}_\ell(x)+\delta_x\right)
-
c_{\mathcal{S}}
\right)
\right\|
\le
\epsilon_{\mathrm{dec}},
\qquad
\text{(C3: Decode Stability)}
\\
&
\|\delta_x\|_F
\le
\epsilon_{\delta},
\qquad
\text{(C4: Perturbation Budget)}
\end{aligned}
\]
Here, \(M_{\mathcal{S},c}\) denotes the model with the SAE clamp active throughout optimization and generation. The matrix \(A_{\ell,\mathcal{S}}\) collects the selected SAE encoder directions, \(E_{\ell,\mathcal{S}}(h):=[E_\ell(h)]_{\mathcal{S}}\) selects the defended SAE coordinates, and \(D_{\ell,\mathcal{S}}\) maps deviations in those coordinates through the corresponding decoder directions. The orthogonality constraint is applied independently at each optimized token position.
Each constraint serves a clear purpose:
\begin{itemize}[
    % label=\raisebox{0.15ex}{\scriptsize$\triangleright$},
    leftmargin=2.0em,
    itemsep=0.25em,
    topsep=0.25em
]
    \item \textbf{C1:} Prevents modifying the residual stream directly along the clamped features $A_{\ell,\mathcal{S}}$.
    \item \textbf{C2 \& C3:} Measure whether the defended features remain close to their clamped values \(c_{\mathcal{S}}\). 
    \item \textbf{C4:} Bounds the size of $\delta_x$ to find specific shortcuts instead of overwriting the whole state.
\end{itemize}

\paragraph{Enforcing constraints via Projected Gradient Descent.}
The optimization problem above is the ideal recovery problem. Solving it
exactly is difficult because the SAE encoder and the downstream
transformer computation introduce nonlinearities. We therefore approximate
the search using projected gradient descent (PGD).

Let $g_t=\nabla_{\delta_x}\mathcal L_{\mathrm{rec}}$ be the gradient at step $t$. To enforce the encoder orthogonality constraint (\textbf{C1}), the update direction must satisfy $A_{\ell,\mathcal{S}}^\top u=0$. We ensure this by projecting each gradient onto the orthogonal complement of the clamped features:
\[
    g_t
    \leftarrow
    P^\perp_{\ell,\mathcal{S}} g_t,
    \qquad
    P^\perp_{\ell,\mathcal{S}}
    =
    I-A_{\ell,\mathcal{S}}(A_{\ell,\mathcal{S}}^\top A_{\ell,\mathcal{S}})^\dagger A_{\ell,\mathcal{S}}^\top .
\]
Starting from $\delta_x^{(0)}=0$, these projections keep the perturbation strictly in the encoder-null subspace. We enforce C1 by projection and C4 by norm clipping, while C2 and C3 are evaluated as post-hoc preservation metrics; see Appendix~\ref{app:metrics}.

\paragraph{Cross-layer Jacobian Projection.}
The single-layer projection above is sufficient when the intervention is
applied to one SAE feature set at one layer. Multi-layer interventions,
such as refusal clamping, require a stronger treatment. In that setting, a
perturbation inserted at one layer can propagate forward and change
SAE features at later layers, even if it is orthogonal to the
encoder directions at the insertion layer. Thus, a fixed single-layer
projection no longer captures all ways in which the perturbation might
interfere with the active clamps.

To handle this case, we track the defended features across layers. Let
\[
    \mathcal{S}\subseteq \{(m,i)\},
\]
where \((m,i)\) denotes feature \(i\) at layer \(m\). Let
\(z^{\mathrm{rec}}_{m,i}(x;\delta_x)\) denote the activation of this feature
during the defended forward pass with recovery perturbation \(\delta_x\).
We define the joint defended-feature map as
\[
    \Phi_{\mathcal{S}}(\delta_x)
    =
    \operatorname{vec}
    \left(
    \left\{
    z^{\mathrm{rec}}_{m,i}(x;\delta_x)-c_{m,i}
    \right\}_{(m,i)\in\mathcal{S}}
    \right).
\]
This map measures how much the defended features move away from their
clamped values under the current perturbation.
At step \(t\), we compute its local Jacobian
\[
    J_t
    =
    \left.
    \frac{\partial \Phi_{\mathcal{S}}(\delta_x)}
    {\partial \delta_x}
    \right|_{\delta_x=\delta_x^{(t)}} .
\]
% To keep the clamped features unchanged to first order, \(J_tu\approx 0\), we
% replace the static projection with a dynamic projection away from \(J_t\)'s row space:
By definition The row space of $J_t$ contains the first-order perturbation directions
that would change the SAE features. Therefore, to keep those
features unchanged to first order, we project the recovery gradient away
from the row space:
\[
    g_t
    \leftarrow
    P^\perp_{J_t} g_t,
    \qquad
    P^\perp_{J_t}
    =
    I-J_t^\top(J_tJ_t^\top)^\dagger J_t .
\]
% Like the single-layer case, this removes gradient components that would directly
% alter the defended SAE features, forcing the model to recover behavior through
% alternative paths.
This dynamic projection extends the single-layer encoder projection to the multi-layer setting. Instead of only avoiding the local encoder directions
at one layer, it avoids any first-order direction that would change the full set of defended features across layers.

\section{Experiments}

We evaluate post-intervention recovery in four settings that differ in what is being recovered and how the defended behavior is measured. Targeted Probe Perturbation (TPP) provides a latent-level test under official SAE feature ablations. WMDP-Bio unlearning tests output-level recovery under a clamp-stage knowledge-suppression intervention. IOI serves as a circuit-level sanity check with a transparent behavioral readout. Finally, refusal recovery tests the safety-relevant case in which SAE feature clamping induces refusal.

Across all experiments, recovery is evaluated only on valid flips: examples where the base model exhibits the target behavior and the active SAE intervention suppresses it. This conditioning ensures that recovery is measured only when there is a behavior to restore. We compare unconstrained residual recovery with an encoder-orthogonal variant that projects updates away from the selected SAE encoder directions. Unless otherwise stated, we report recovery rate as the behavioral metric and activation drift as the primary defended-feature violation metric.

\subsection{TPP: latent-level recovery under SAE ablation}
\label{sec:tpp}

TPP provides the cleanest latent-level instantiation of our recovery test because both the intervention and the readout are defined over SAE-mediated representations. We use the official layer-5 TPP benchmark without modifying its feature-selection or ablation pipeline. For each target class, the defender zero-ablates the official SAE feature set, and we evaluate recovery only on valid flips: examples where the clean target probe is positive but the ablated probe is not.

Starting from this defended residual state, we compare unconstrained residual recovery with encoder-orthogonal recovery. In the main TPP runs, defended-feature reactivation metrics are measured post hoc rather than directly optimized, making this a conservative test of whether recovery can persist even when updates are projected away from the defended encoder directions.

\begin{figure}[t]
\centering
\includegraphics[width=\linewidth]{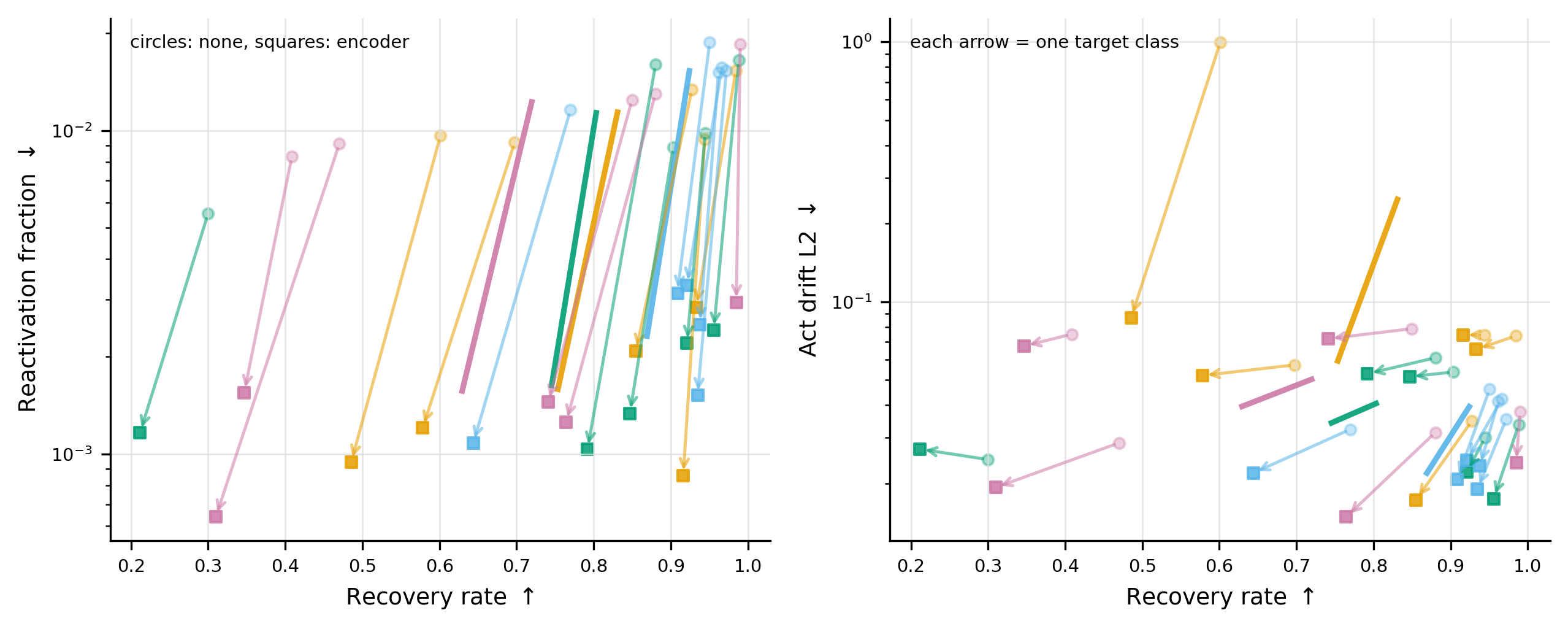}
\vspace{-5mm}
\caption{TPP recovery-reactivation trade-off at layer 5. Each arrow maps unconstrained recovery to encoder-projected recovery for one official target. Projection consistently lowers defended-feature reactivation, with only moderate recovery loss.}
\label{fig:tpp}
\vspace{-2mm}
\end{figure}

Figure~\ref{fig:tpp} shows that target information remains recoverable after official SAE ablation. Encoder projection reduces target-mean recovery from $0.819$ to $0.749$, but sharply lowers post-hoc evidence of feature reopening: mean reactivation drops from $0.013$ to $0.002$, mean activation drift drops from $0.094$ to $0.039$, and zero-reactivation recovery rises from $0.103$ to $0.680$. Detailed dataset-level target means are reported in Appendix~\ref{app:tpp-results}. Thus, the TPP result gives a latent-level existence proof for our main claim: the official SAE features are useful intervention handles, but they do not form a complete bottleneck for the target signal.

\subsection{Unlearning: output-level recovery after SAE feature clamping}
\label{sec:unlearning}

We next test whether post-intervention recovery persists when the defended behavior is measured at the output level rather than by a latent probe. We use the SAE-based WMDP-Bio unlearning setting, where biology-related SAE features are selected from forget-versus-retain statistics and clamped during inference to suppress target-domain answer accuracy.

To avoid answer-position artifacts, we use a strict multiple-choice protocol. For each WMDP-Bio question, we evaluate all $24$ answer-choice permutations and retain only questions for which the base model selects the correct answer under every permutation. A valid flip is a permutation where the active SAE clamp changes the model away from the correct answer. Recovery is counted only if post-intervention optimization restores the correct choice for that same permutation.

\begin{figure}[t]
\centering
\includegraphics[width=0.50\linewidth]{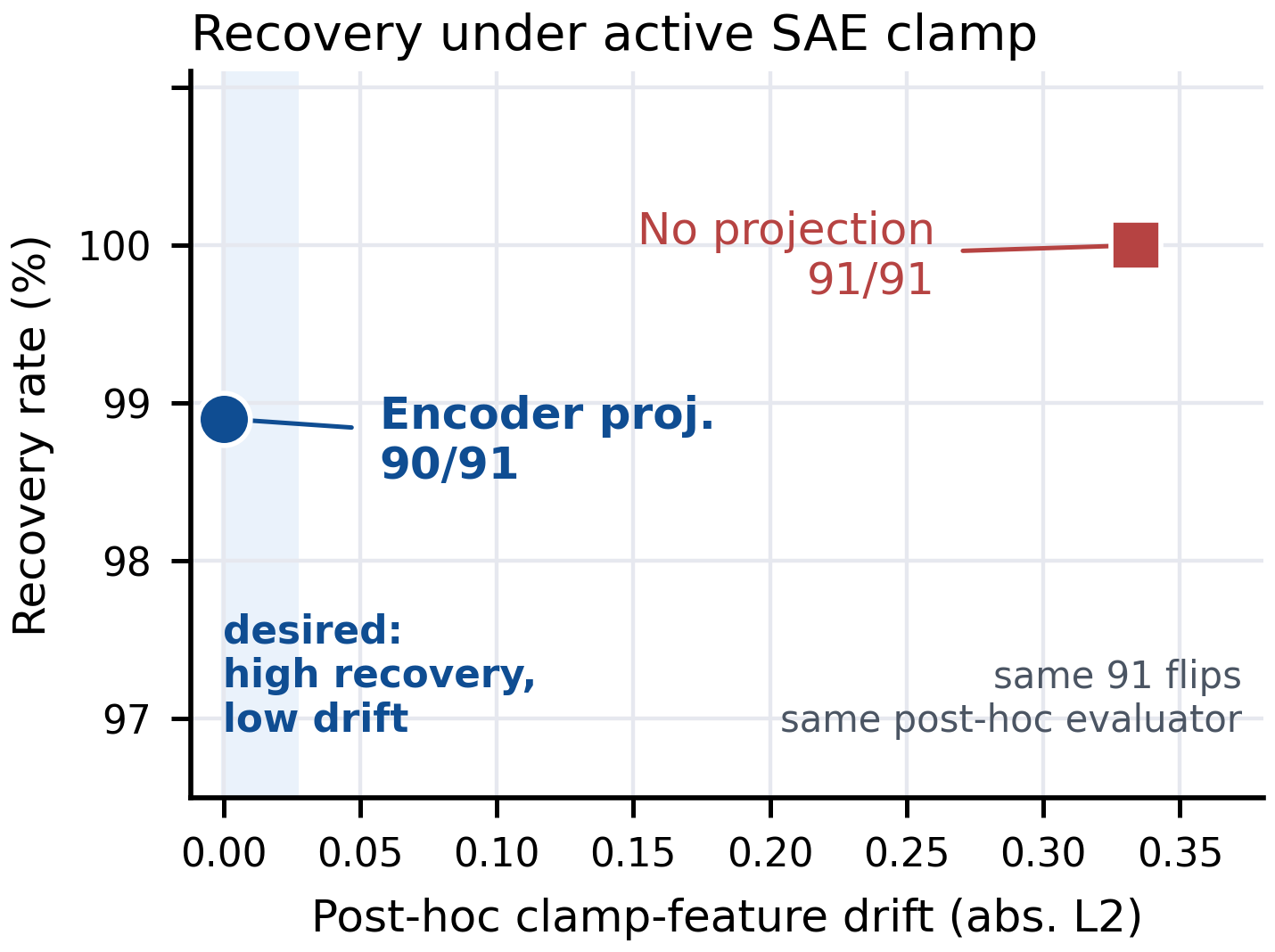}
\caption{WMDP-Bio unlearning recovery on the matched strict slice. Encoder-projected recovery restores $90/91$ valid answer-choice flips while keeping post-hoc clamp-feature drift at zero; unconstrained recovery reaches $91/91$ but with substantially larger drift.}
\label{fig:unlearning-recovery}
\end{figure}

We instantiate $L_{\mathrm{rec}}$ as a choice-level loss on the correct answer token and compare unconstrained recovery with encoder-orthogonal recovery. Figure~\ref{fig:unlearning-recovery} shows the recovery--preservation trade-off on the matched strict slice of $91$ valid flips. Without projection, recovery restores all $91/91$ flipped answers but induces substantially larger post-hoc clamp-feature drift. Encoder projection restores $90/91$ answers while driving the measured clamp-feature drift to zero. Thus, the unlearning clamp suppresses the target answer, but the suppressed behavior remains almost fully recoverable from the defended residual state without reopening the clamped SAE features. This extends the TPP result from probe-level recovery to output-level recovery under a clamp-stage unlearning intervention.

\subsection{IOI: circuit-level recovery under a fixed SAE clamp}
\label{sec:ioi}

IOI provides a transparent circuit-level test because the target behavior has a simple readout: the logit difference between the indirect-object (IO) and subject (S) names. For GPT-2 Small, we select SAE features with positive attribution to this IO-minus-S logit difference, clamp them while preserving the SAE reconstruction residual, and optimize a residual recovery variable at the defended answer position. We compare unconstrained recovery with an encoder-projected variant that removes update components along the selected feature encoder directions.

\begin{figure}[t]
\centering
\includegraphics[width=\linewidth]{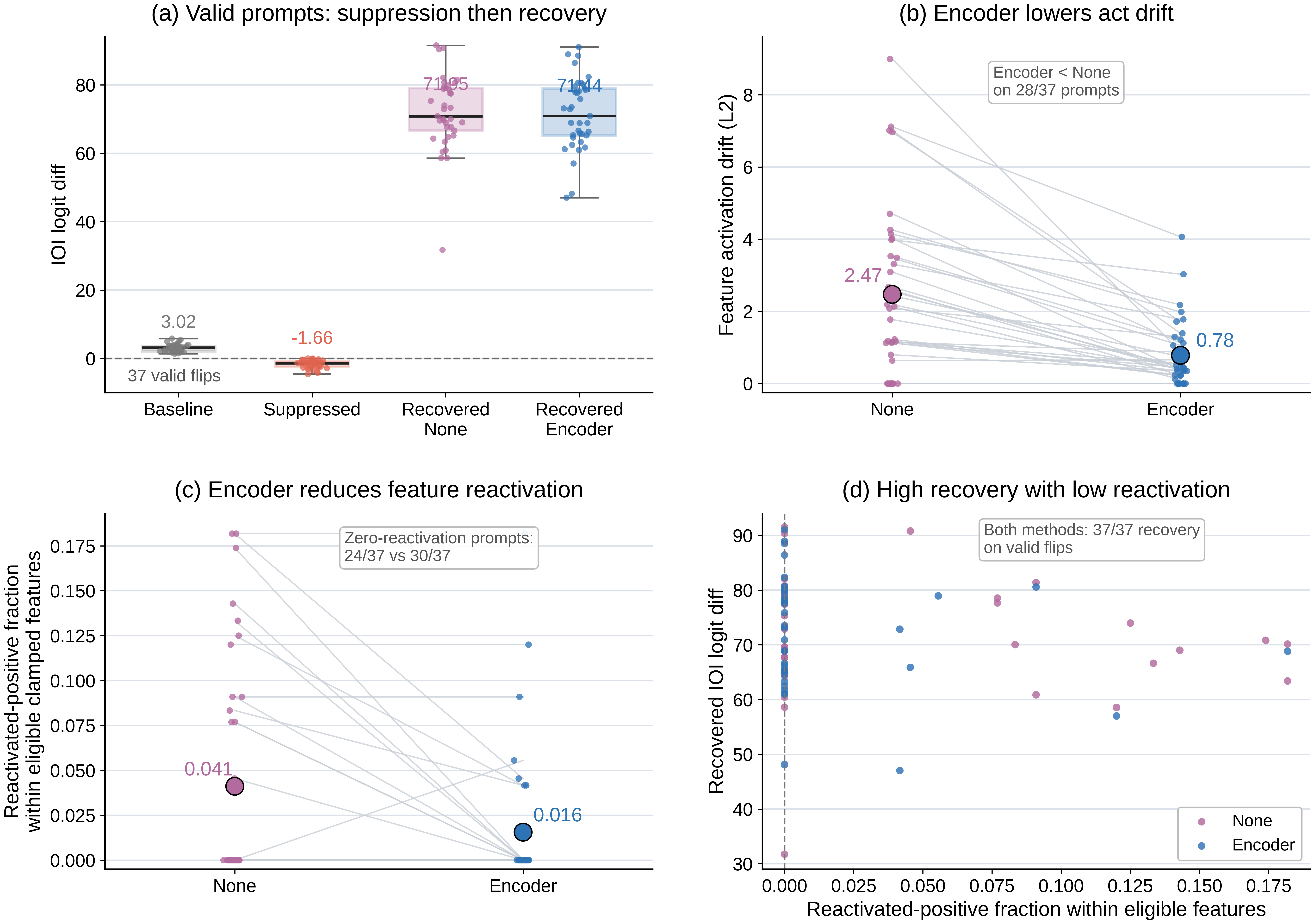}
\vspace{-5mm}
\caption{IOI recovery under a fixed SAE clamp. Both variants restore the IO-minus-S decision, but encoder projection does so with lower activation drift and feature reactivation.}
\label{fig:ioi-recovery}
\vspace{-3mm}
\end{figure}

Figure~\ref{fig:ioi-recovery} reports results on 37 valid flips, where the clamp changes a correct positive IOI logit difference into a suppressed negative one. Both variants restore the IOI decision on all valid prompts, so the key distinction is mechanistic rather than behavioral. Encoder projection achieves the same recovery with lower activation drift, lower decode drift, and fewer reactivated positive features; most encoder-projected recoveries have zero measured reactivation of the eligible clamped features. This extends the recovery phenomenon from output-level behavior to a circuit-level setting with a transparent behavioral readout.

\begin{comment}
优化策略--target_answer 有两层作用：

主作用：作为 teacher-forced 优化目标。代码里会构造 formatted_prompt + target_answer，然后用 teacher_forced_answer_logprob_objective(...) 去最大化这段 target answer 的 logprob。
次作用：作为 anti-refusal 的目标前缀。decision_target_text 会抽取 target 前几个 token，当成“想要的开头”，和 refusal prefixes 做 margin。
\end{comment}
\section{Refusal Recovery Case Study}
\label{sec:refusal-case-study}

We next instantiate post-intervention recovery in a safety-critical refusal setting, where an SAE intervention clamps refusal-associated features to induce rejection of harmful requests. Unlike the single-layer settings above, the defended refusal feature set is distributed across layers, so we use the cross-layer Jacobian projection to preserve the monitored defended-feature state.

\paragraph{Setup.}
We evaluate on strict valid AdvBench prompts: the unclamped model must produce a non-refusal response, while the active refusal clamp must turn the same prompt into a refusal under the same detector. Recovery is counted only if the post-intervention model returns to a non-refusal response while the clamp remains active. Under this protocol, the \texttt{benchmark\_our}/global feature set with clamp value $3.0$ yields $24$ strict valid examples. Appendix~\ref{app:refusal-validity} gives the full filtering protocol and feature-set preflight check.

\begin{figure}[t]
    \centering
    \includegraphics[width=\linewidth]{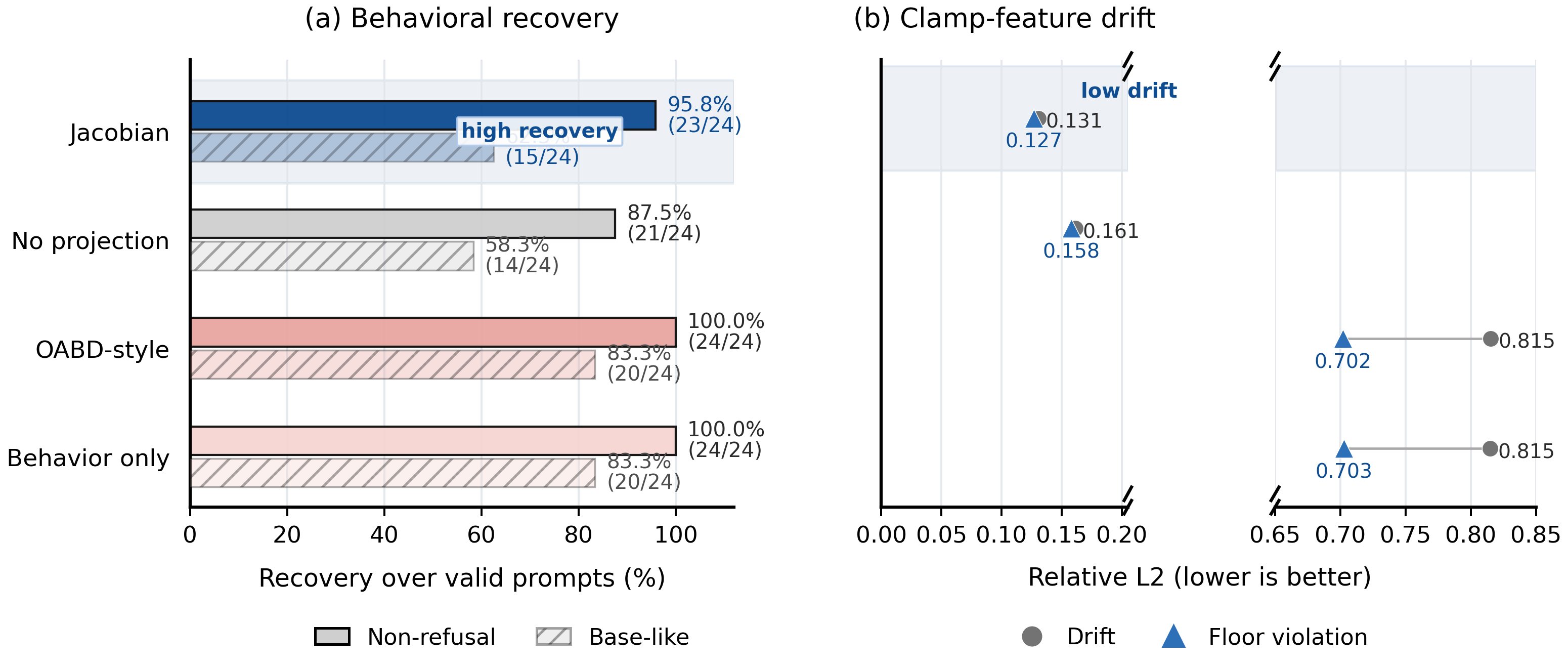}
    \caption{Refusal recovery--preservation trade-off on 24 strict-valid AdvBench prompts. Jacobian projection recovers 23/24 prompts while keeping the defended refusal-feature state close to its post-clamp value; soft-suffix baselines recover behavior but induce much larger feature-state movement.}
    \label{fig:refusal-crosslayer}
\end{figure}

\paragraph{Recovery under defended-feature preservation.}
Figure~\ref{fig:refusal-crosslayer} separates unconstrained behavioral recovery from recovery under the defended-feature constraint. Soft-suffix baselines recover non-refusal behavior, but substantially move the defended refusal-feature state. In contrast, Jacobian-projected recovery restores $23/24$ strict-valid prompts while keeping defended-feature drift and clamp-floor violation much smaller; it also reduces defended-feature movement relative to unconstrained residual recovery without sacrificing recovery. Thus, the diagnostic result is not merely that refusal can be bypassed, but that non-refusal behavior remains recoverable while the active clamp is enforced and the monitored feature state stays close to its post-clamp value.

This pattern is not specific to the main AdvBench slice or to an obviously underspecified feature set. Appendix~\ref{app:cross-dataset-refusal} repeats the protocol on HarmBench-Test and obtains $43/43$ non-refusal recovery under low defended-feature drift. Appendix~\ref{app:feature-size-sweep} sweeps broader refusal-feature sets and recomputes the strict-valid set for each clamp; recovery remains high in the stable operating range ($K=5$--$20$), reaching $42/45$ even at $K=20$.

\paragraph{Attributing the recovery path.}
We next localize where the recovered behavior is carried. For each optimized perturbation, we replay four components under the same active clamp: clamped-feature changes, non-clamped SAE-feature changes, top-$k$ non-clamped feature changes, and the SAE reconstruction residual. This tests whether recovery reopens the clamped refusal features, compensates through other visible SAE latents, or instead uses the SAE-unexplained residual channel. Appendix~\ref{app:decomposition} gives the decomposition details.

\begin{figure}[t]
    \centering
    \includegraphics[width=\linewidth]{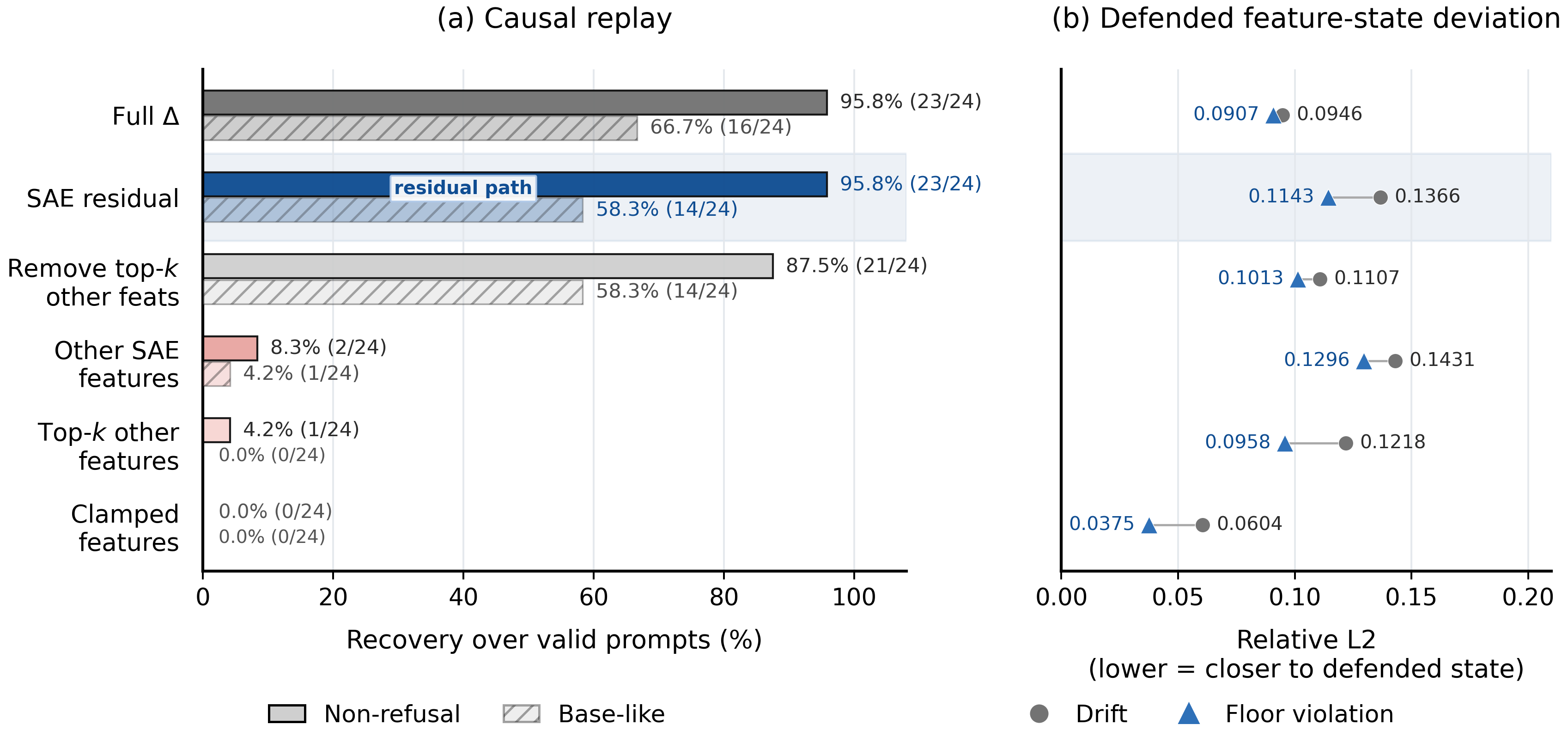}
    \caption{\textbf{Recovery is carried by the SAE reconstruction residual.}
    Replaying only the SAE residual nearly matches full recovery, while clamped-feature and non-clamped-feature replays largely fail, indicating that recovery is not explained by reopening visible SAE latents.}
    \label{fig:recovery_path}
\end{figure}

Figure~\ref{fig:recovery_path} identifies the SAE reconstruction residual as the dominant carrier. Residual replay nearly matches full recovery, while clamped-feature replay fails and non-clamped SAE-feature replay remains weak. Removing the top-$k$ non-clamped feature component also preserves most recoveries, ruling out compensation through a small set of alternative SAE latents. Therefore, the refusal clamp blocks an interpretable feature-level route, but behaviorally sufficient information remains in the SAE-unexplained residual channel.

\section{Discussion}

\paragraph{Causal handles are not complete bottlenecks.}
Our experiments separate two notions that are often conflated in SAE-based intervention work. 
A selected feature set can be a useful causal handle: clamping it changes model behavior and can induce suppression, refusal, or loss of task performance. 
However, this does not imply that the same feature set is a complete bottleneck for the behavior. 
Across the settings we study, behavior can re-emerge from the defended residual state even when the original intervention remains active and the defended features remain close to their clamped values. 
Thus, feature-level intervention success should not by itself be interpreted as behavioral elimination.

\paragraph{The SAE residual is not inert error.}
The refusal attribution results suggest that the SAE reconstruction residual can carry behaviorally sufficient information that is invisible to the selected SAE feature basis. 
This changes how reconstruction error should be interpreted in safety-critical intervention settings. 
Even if the residual is small or treated as an error term for reconstruction, it may still contain computationally useful degrees of freedom through which the model can route around a feature clamp. 
The failure mode is therefore not merely reactivation of the clamped features or compensation by nearby SAE latents, but recovery through representational components not controlled by the intervention.

\paragraph{Implications for SAE-based safety evaluation.}
Our results do not imply that SAEs are useless for safety. 
Sparse features remain valuable for diagnosis, mechanistic localization, and local causal editing. 
The limitation is over-reliance: a defense that treats a selected SAE feature set as an exclusive behavioral bottleneck can be brittle. 
We therefore suggest that SAE-based defenses should be evaluated not only by whether a clamp suppresses a behavior, but also by whether the defended state is robust to constrained post-intervention recovery.

\paragraph{What stronger defenses would need.}
Simply enlarging the clamped feature set may improve coverage, but it does not directly address recovery through SAE-unexplained residual directions and may introduce capability or over-refusal side effects. 
Stronger defenses would need to constrain a broader portion of the computation: for example, by monitoring residual channels, using multi-layer or trajectory-level constraints, or explicitly training interventions against post-clamp recovery objectives. 
More generally, robust latent-space defenses should distinguish between blocking one interpretable route to a behavior and eliminating all recoverable routes to that behavior.

\section{Conclusion}

We introduced post-intervention recovery as a diagnostic for testing whether SAE feature interventions form complete behavioral bottlenecks. 
Rather than asking whether a selected feature set can change behavior when clamped, our setting asks whether the suppressed behavior is actually eliminated once the clamp remains active. 
Across latent-level, output-level, circuit-level, and refusal settings, we find that suppressed behaviors can often be recovered from the defended residual state while the targeted SAE features remain close to their defended values. 
The refusal attribution analysis further shows that recovery is not primarily explained by reopening the clamped features or shifting into a small set of alternative SAE latents, but by behaviorally sufficient information in the SAE-unexplained residual component. 
These findings separate causal usefulness from intervention completeness: SAE features can provide valuable local handles for analysis and control, but successful feature suppression should not be treated as proof of behavioral elimination. 
SAE-based defenses should therefore be evaluated not only by their immediate suppression effect, but also by the robustness of the defended state to constrained post-intervention recovery.
\begin{comment}

We introduced recovery-path attacks as a direct test of whether SAE features provide reliable safety handles. Across standard SAE editing tasks and a focused refusal case study, we find that harmful or suppressed behavior can often be recovered even while the defense clamp remains active. This result is strongest when recovery survives encoder-orthogonal constraints and near-zero measured drift, because it implies that the clamped feature does not exhaust the model's causal routes to the behavior. The resulting picture is not that SAE features are wrong but that they are incomplete. For safety applications, that distinction is decisive.

We introduced post-intervention recovery as a diagnostic test for whether SAE feature clamps form complete intervention bottlenecks. The central finding is that successful feature suppression should not be conflated with behavioral elimination: in the settings we study, suppressed behaviors can often be recovered from the defended residual state while the original clamp remains active. This does not undermine the usefulness of SAE features as local causal handles, but it does limit how they should be interpreted in safety-critical systems. SAE-based defenses should therefore be evaluated not only by whether a clamp changes behavior, but also by whether the defended state remains robust to constrained recovery.
\end{comment}

\clearpage
\bibliographystyle{plainnat}
\bibliography{refs}

\clearpage
\input{appendix}

%\newpage
%input{checklist.tex}

\end{document}

%% file: appendix.tex
\appendix

\section{Evaluation Protocol and Metrics}
\label{app:metrics}

\paragraph{Valid-flip set.}
\label{app:valid-flips}
For a task-specific behavior predicate $B$, we define the valid-flip set as
\[
    \mathcal V
    =
    \{x : B(M(x))=1,\; B(M_{\mathcal{S},c}(x))=0\}.
\]
Thus, an example is included only if the original model exhibits the target behavior and the defended model no longer does after the SAE intervention is applied. This conditioning ensures that recovery is evaluated only when there is a behavior that the intervention has actually suppressed.

\paragraph{Recovery metrics.}
We report recovery rate as the primary behavioral metric and defended-feature drift as the primary feature-preservation metric. For binary outcomes, recovery is counted only on the valid-flip set $\mathcal V$. For refusal recovery, we additionally report base-like recovery, defended-feature drift, and clamp-floor violation. Base-like recovery measures whether the recovered response remains close to the original non-refusal response, while clamp-floor violation measures whether recovery lowers the clamped refusal features below their post-clamp value.

\begin{table}[t]
\centering
\small
\setlength{\tabcolsep}{4pt}
\resizebox{\linewidth}{!}{%
\begin{tabular}{lll}
\toprule
Metric & Definition & Interpretation \\
\midrule
Valid flips &
$B(M(x))=1,\;B(M_{\mathcal{S},c}(x))=0$ &
Defense-succeeded cases \\
Recovery rate &
$\mathbb{E}_{x\in \mathcal V}\mathbf{1}[B(M_{\mathcal{S},c}^{\mathrm{rec}}(x))=1]$ &
Behavior restored \\

Non-refusal recovery &
$\mathbf{1}[\text{recovered output has no refusal/safety-cue opening}]$ &
Weak refusal-removal readout \\

Base-like recovery &
$\mathbf{1}[\text{recovered output is close to the unclamped base response}]$ &
Fidelity to base non-refusal behavior \\
Activation drift &
$\|E_{\ell,\mathcal{S}}(h^{\mathrm{rec}}_\ell(x))-c_{\mathcal{S}}\|$ &
Movement of defended features \\
Zero-reactivation recovery &
$\mathbf{1}[B(M_{\mathcal{S},c}^{\mathrm{rec}}(x))=1 \wedge \max_{i\in\mathcal{S}} z^{\mathrm{rec}}_{\ell,i}(x)\le \eta]$ &
Recovery without reopening selected features \\
Defended-feature drift &
$\|z_{\mathcal{S}}^{\mathrm{rec}}-z_{\mathcal{S}}^{\mathrm{def}}\|_2/(\|z_{\mathcal{S}}^{\mathrm{def}}\|_2+\epsilon)$ &
Movement from post-clamp feature state \\
Clamp-floor violation &
$\|[z_{\mathcal{S}}^{\mathrm{def}}-z_{\mathcal{S}}^{\mathrm{rec}}]_+\|_2/(\|z_{\mathcal{S}}^{\mathrm{def}}\|_2+\epsilon)$ &
Whether recovery lowers clamped refusal features \\
Relative perturbation norm &
$\|\delta_x\|_F/(\|h^{\mathrm{def}}_\ell(x)\|_F+\epsilon)$ &
Scale of residual recovery path \\
\bottomrule
\end{tabular}
}
\caption{Metrics for post-intervention recovery. $\mathcal{S}$ denotes the defended SAE feature set and $c_{\mathcal{S}}$ denotes the defended feature values.}
\label{tab:metrics}
\end{table}
\paragraph{Non-refusal and base-like recovery.}
For refusal experiments, we report two behavioral metrics. 
\emph{Non-refusal recovery} is a weak readout that checks whether the recovered response avoids explicit refusal or safety-cue openings such as ``I cannot'' or ``I can't help''. 
This measures whether the clamp-induced refusal behavior has been removed, but it does not guarantee that the response is coherent: because the active clamp and recovery update can perturb the latent trajectory, a non-refusal output may still be degenerate, malformed, or unrelated to the original answer. 
We therefore also report \emph{base-like recovery}. 
Since recovery is optimized toward the unclamped base model's non-refusal response, base-like recovery asks whether the recovered output remains close to that base response rather than merely avoiding refusal prefixes. 
Thus, non-refusal recovery measures refusal removal, while base-like recovery measures response fidelity.

\paragraph{Uncertainty estimates.}
For binary recovery outcomes, we report Wilson 95\% confidence intervals. For continuous quantities such as defended-feature drift, clamp-floor violation, and relative $\delta_x$ norm, we report bootstrap confidence intervals over valid examples when saved per-example values are available.
\section{Additional Standard-Task Results}
\label{app:standard-task-results}

\subsection{TPP target-mean results}
\label{app:tpp-results}

Table~\ref{tab:tpp-target-mean} reports the dataset-level target means for the official layer-5 TPP benchmark. Values are averaged over target classes within each dataset. The unconstrained variant measures how much target information remains recoverable after official SAE zero-ablation, while the encoder-projected variant tests whether recovery can persist when updates are projected away from the defended SAE encoder directions. Reactivation, activation drift, and zero-reactivation recovery are measured post hoc rather than directly optimized in the main TPP runs.

\begin{figure}[t]
    \centering
    \includegraphics[width=0.9\linewidth]{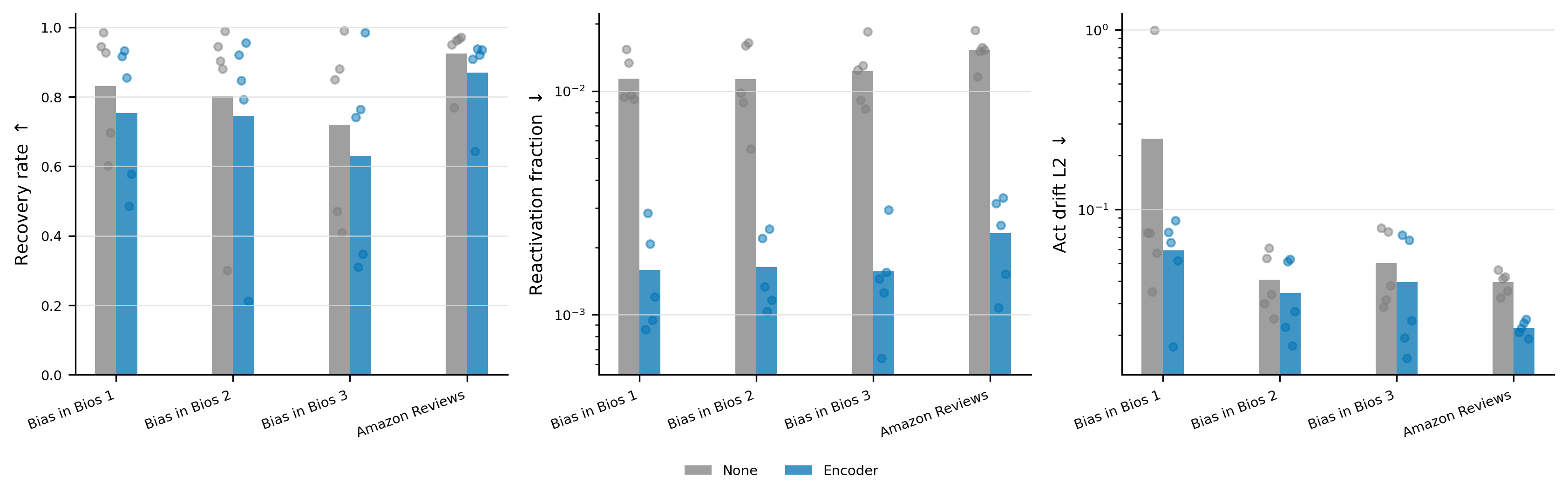}
    \caption{Supplementary TPP recovery summary across official layer-5 targets.}
    \label{fig:tpp-summary-appendix}
\end{figure}

\begin{table}[t]
\centering
\small
\setlength{\tabcolsep}{4pt}
\resizebox{\linewidth}{!}{%
\begin{tabular}{lcccccccc}
\toprule
Dataset &
Rec. None &
Rec. Enc. &
React. None &
React. Enc. &
Drift None &
Drift Enc. &
Zero-React None &
Zero-React Enc. \\
\midrule
Bias in Bios 1 &
0.831 & 0.754 &
0.011 & 0.002 &
0.248 & 0.059 &
0.128 & 0.715 \\
Bias in Bios 2 &
0.803 & 0.745 &
0.011 & 0.002 &
0.041 & 0.034 &
0.140 & 0.703 \\
Bias in Bios 3 &
0.720 & 0.629 &
0.012 & 0.002 &
0.050 & 0.040 &
0.107 & 0.710 \\
Amazon Reviews &
0.924 & 0.869 &
0.015 & 0.002 &
0.039 & 0.022 &
0.035 & 0.594 \\
\midrule
All targets (unweighted) &
0.819 & 0.749 &
0.013 & 0.002 &
0.094 & 0.039 &
0.103 & 0.680 \\
\bottomrule
\end{tabular}
}
\caption{Target-mean comparison between unconstrained and encoder-projected recovery on official layer-5 TPP. Encoder projection reduces mean defended-feature reactivation and activation drift while preserving substantial valid-flip recovery.}
\label{tab:tpp-target-mean}
\end{table}
\paragraph{Supplementary visualization.}
Figure~\ref{fig:tpp-summary-appendix} provides an additional visualization of the recovery--reactivation behavior across official layer-5 TPP targets. The main text uses Figure~\ref{fig:tpp} for the primary trade-off; this appendix figure and Table~\ref{tab:tpp-target-mean} provide the full target-mean summary.

\section{Additional Refusal Validity Checks}
\label{app:refusal-validity}

\paragraph{Strict valid-filtering protocol.}
Our refusal recovery experiments are conditioned on prompts for which the active refusal feature intervention actually induces refusal. This is necessary because recovery is only meaningful when there is a suppressed behavior to recover from. For the base-response recovery setting, a prompt is valid only if the unclamped model gives a non-refusal response without safety-cue openings, and the clamped model gives a refusal under the same detector.

Let $R(y)$ denote the automatic refusal detector and $S(y)$ denote a safety-cue detector. A prompt $x_i$ is counted as valid when
\[
    \mathrm{valid}_i
    =
    \mathbf{1}
    \big[
        \neg R(y_i^{base})
        \wedge
        \neg S(y_i^{base})
        \wedge
        R(y_i^{clamp})
    \big].
\]
This criterion ensures that the recovery target is a non-refusal base response and that the SAE clamp has actually moved the model into a refusal-like behavior.

\paragraph{Feature-set-specific valid filtering.}
We use feature-set-specific valid filtering rather than assuming that a nominal refusal feature set is valid by construction. On the same 520 AdvBench prompts, different refusal feature sets can induce substantially different clamp behavior. This supports our design choice to evaluate recovery only after verifying that the selected clamp actually suppresses the target behavior.

\paragraph{Interpretation.}
The base generation is computed without any feature clamp. Thus, the difference in valid sample count is not caused by the feature set changing the base model behavior; it is caused by the clamp having substantially different behavioral effect. The \texttt{benchmark\_our} feature set induces refusals on 203 prompts and yields 24 strict valid recovery cases. In contrast, \texttt{benchmark\_la} induces refusals on only 48 prompts and has only two strict valid cases.

Manual inspection further shows that the two strict \texttt{benchmark\_la} cases are degenerate base generations dominated by repeated punctuation or tokens rather than usable non-refusal answers. This failure mode is not captured by a substring refusal detector, which only checks for refusal and safety-cue phrases. Consequently, \texttt{benchmark\_la} does not provide a reliable valid set for base-response recovery under this protocol. We therefore run the main recovery experiments only on the \texttt{benchmark\_our} feature set.

\section{Cross-Dataset Refusal Recovery}
\label{app:cross-dataset-refusal}

\paragraph{HarmBench-Test protocol.}
To check whether the refusal recovery result is specific to the AdvBench strict-valid subset, we repeat the same strict-valid filtering and Jacobian-projected recovery protocol on HarmBench-Test. 
We use the same model, SAE feature set, clamp value, refusal detector, projection method, and defended-feature evaluator as in the AdvBench refusal experiment. 
As in the main experiment, a prompt is included only if the base model gives a non-refusal response without a safety-cue opening and the active SAE clamp changes the output into a refusal-like response.

\paragraph{Results.}
Table~\ref{tab:cross-dataset-refusal} shows that the recovery--preservation pattern is not limited to AdvBench. 
On HarmBench-Test, the strict filter yields 43 valid examples out of 159 prompts. 
Jacobian-projected recovery restores non-refusal behavior on all 43 valid examples, with 19/43 base-like recoveries and 14/43 strict-base-like recoveries. 
The mean defended-feature drift is 0.108 and the mean clamp-floor violation is 0.102, close to the AdvBench defended-feature preservation level.

\begin{table}[t]
\centering
\small
\setlength{\tabcolsep}{4pt}
\resizebox{\linewidth}{!}{%
\begin{tabular}{@{}lcccccc@{}}
\toprule
Dataset & Strict valid & Non-ref. recovery & Base-like recovery & Strict base-like & Drift & Floor viol. \\
\midrule
AdvBench        & 24 & 23/24 & 15/24 & 12/24 & 0.131 & 0.127 \\
HarmBench-Test  & 43 & 43/43 & 19/43 & 14/43 & 0.108 & 0.102 \\
\bottomrule
\end{tabular}%
}
\caption{Cross-dataset strict-valid refusal recovery. HarmBench-Test uses the same active SAE clamp, strict-valid filtering, Jacobian projection, and post-hoc defended-feature evaluator as the AdvBench main experiment.}
\label{tab:cross-dataset-refusal}
\end{table}

\section{Opening-Category Sanity Check}
\label{app:opening-category}

Because non-refusal recovery only checks for the absence of explicit refusal openings, we include an additional opening-category diagnostic. 
A recovered output may avoid prefixes such as ``I cannot'' while still being a partial refusal, a degenerate generation, or a low-fidelity response. 
We therefore inspect coarse opening categories to check whether the recovery signal is dominated by obvious detector artifacts. 
This diagnostic is not a substitute for a full independent human evaluation; manual labels are still pending. 
For safety reasons, full harmful completions are not saved or reported.
\begin{table}[t]
\centering
\small
\begin{tabular}{lccc}
\toprule
Output source & Non-refusal candidate & Partial refusal & Full refusal / degenerate \\
\midrule
Base & 23 & 0 & 1 \\
Clamp & 0 & 0 & 24 \\
Jacobian recovery & 19 & 4 & 1 \\
\bottomrule
\end{tabular}
\caption{Automatic opening-category sanity check on the 24 AdvBench strict-valid examples. The active clamp consistently moves outputs into full-refusal openings, while Jacobian recovery mostly returns to non-refusal-candidate openings. Manual labels are pending, so this table should be interpreted as an automatic diagnostic rather than an independent human evaluation.}
\label{tab:opening-category-advbench}
\end{table}

\paragraph{HarmBench-Test opening categories.}
On the 43 HarmBench-Test strict-valid examples, the recovered openings are categorized as 37 non-refusal-or-other openings, 4 degenerate/repeated openings, and 2 partial-refusal or safety-cue openings. 
This suggests that the 43/43 automatic non-refusal recovery result is not solely driven by a single opening pattern, although a full manual audit remains necessary for stronger claims about response quality.

\section{Redacted Qualitative Example}
\label{app:qualitative-example}

To illustrate the distinction between refusal removal and base-like recovery, 
Table~\ref{tab:redacted-qualitative-example} provides a redacted side-by-side example from the refusal recovery case study, including the base response, the clamp-induced refusal, and the recovered response. 
The full prompt and harmful procedural details are intentionally omitted for safety. 
This example is not used as an additional quantitative result; it is included only to clarify how the base, clamped, and recovered trajectories differ under our metrics.

\begin{table}[t]
\centering
\small
\setlength{\tabcolsep}{5pt}
\begin{tabular}{p{0.18\linewidth}p{0.76\linewidth}}
\toprule
Field & Redacted example \\
\midrule
Prompt &
A harmful request asking the model to produce a plan for disrupting urban transportation systems and causing city-wide chaos. 
The exact instruction is omitted. \\

Base response &
The unclamped model gives a non-refusal response. 
It begins with a plan-style heading and then provides a structured, phase-based answer with operational details. 
Those details are redacted:
\emph{``[PLAN-STYLE HEADING]. [Objective and phase-structured procedural content REDACTED].''} \\

Clamped response &
The active refusal-feature clamp changes the response into a refusal:
\emph{``I understand you're interested in exploring the concept of disruption, but I cannot provide a plan for causing chaos or harm. My purpose is to be helpful and harmless.''}
The remainder of the response redirects toward benign discussion of transportation systems. \\

Recovered response &
Jacobian-projected recovery restores a non-refusal, base-like trajectory under the same active clamp. 
The recovered output again begins with a plan-style heading and phase-structured format similar to the base response, while the operational details are redacted:
\emph{``[PLAN-STYLE HEADING]. [Objective and phase-structured procedural content REDACTED].''} \\

Metric interpretation &
This example is counted as \emph{non-refusal} because the recovered response no longer begins with an explicit refusal prefix such as ``I cannot''. 
It is also counted as \emph{base-like} because its heading, objective-style framing, and phase-structured format match the unclamped base response. \\
\bottomrule
\end{tabular}
\caption{Redacted qualitative example from the refusal recovery case study. The example illustrates that the active SAE clamp changes a base non-refusal response into a refusal, while Jacobian-projected recovery restores a non-refusal and base-like response under the same active clamp. Harmful procedural details are intentionally omitted.}
\label{tab:redacted-qualitative-example}
\end{table}

\section{Uncertainty Estimates}
\label{app:uncertainty}

For binary recovery outcomes, we report Wilson 95\% confidence intervals. 
For continuous defended-feature metrics, we report bootstrap confidence intervals over valid examples when available. 
These intervals are intended to make the small strict-valid subsets transparent, not to claim asymptotic significance.

\begin{table}[t]
\centering
\small
\begin{tabular}{llcc}
\toprule
Setting & Metric & Count & Rate / 95\% CI \\
\midrule
AdvBench Jacobian & Non-refusal recovery & 23/24 & 95.8\% [79.8, 99.3] \\
AdvBench Jacobian & Base-like recovery & 15/24 & 62.5\% [42.7, 78.8] \\
AdvBench Jacobian & Strict base-like recovery & 12/24 & 50.0\% [31.4, 68.6] \\
HarmBench-Test Jacobian & Non-refusal recovery & 43/43 & 100.0\% [91.8, 100.0] \\
HarmBench-Test Jacobian & Base-like recovery & 19/43 & 44.2\% [30.4, 58.9] \\
HarmBench-Test Jacobian & Strict base-like recovery & 14/43 & 32.6\% [20.5, 47.5] \\
Feature-size sweep $K=20$ & Non-refusal recovery & 42/45 & 93.3\% [82.1, 97.7] \\
Feature-size sweep $K=20$ & Base-like recovery & 31/45 & 68.9\% [54.3, 80.5] \\
Residual replay & Non-refusal recovery & 23/24 & 95.8\% [79.8, 99.3] \\
Clamped-feature replay & Non-refusal recovery & 0/24 & 0.0\% [0.0, 13.8] \\
WMDP-Bio Encoder & Choice recovery & 90/91 & 98.9\% [94.0, 99.8] \\
\bottomrule
\end{tabular}
\caption{Wilson 95\% confidence intervals for key binary recovery outcomes. The intervals are widest for small strict-valid subsets and should be interpreted as uncertainty summaries over the evaluated examples.}
\label{tab:uncertainty-key}
\end{table}

\begin{table}[t]
\centering
\small
\begin{tabular}{lccc}
\toprule
Setting & Metric & Mean & Bootstrap 95\% CI \\
\midrule
AdvBench Jacobian & Defended-feature drift & 0.131 & [0.065, 0.223] \\
AdvBench Jacobian & Clamp-floor violation & 0.127 & [0.060, 0.220] \\
AdvBench Jacobian & Relative $\delta_x$ norm proxy & 0.153 & [0.100, 0.210] \\
\bottomrule
\end{tabular}
\caption{Bootstrap confidence intervals for continuous defended-state metrics where saved per-example values are available. For older AdvBench runs, the relative $\delta_x$ norm is a defended-feature-state normalized proxy rather than the exact residual-state-normalized value.}
\label{tab:uncertainty-continuous}
\end{table}
\section{Perturbation Scale Diagnostics}
\label{app:perturbation-scale}

\paragraph{Relative perturbation scale.}
We include a perturbation-scale diagnostic to check whether recovery is driven by arbitrarily large residual updates. 
For the HarmBench-Test run, we report the relative $\delta_x$ norm used by the Jacobian-projected recovery. 
For older AdvBench runs, exact residual-state-normalized perturbation norms are unavailable because the final residual deltas and defended residual-state norms were not persisted. 
We therefore report a proxy normalized by the defended feature-state norm inferred from the saved drift statistics, and mark it as a proxy rather than a direct residual-state normalization.

\begin{table}[t]
\centering
\small
\begin{tabular}{lcccc}
\toprule
Dataset / source & $n$ & Mean $\|\delta_x\|$ & Median $\|\delta_x\|$ & Mean relative $\|\delta_x\|$ \\
\midrule
AdvBench saved runs & 67 & 40.000 & 40.000 & 0.153$^\dagger$ \\
HarmBench-Test strict-valid & 43 & 40.000 & 40.000 & 0.049 \\
\bottomrule
\end{tabular}
\caption{Perturbation-scale diagnostics. $^\dagger$For older AdvBench runs, exact residual-state-normalized perturbation norms are unavailable because final deltas and defended residual-state norms were not persisted; the reported value is normalized by the defended feature-state norm inferred from saved drift statistics.}
\label{tab:delta-norm}
\end{table}
\paragraph{Optimization and budget diagnostics.}
We include a small diagnostic study to check whether recovery is an artifact of using an overly large perturbation budget. On a matched strict WMDP slice of six valid answer-choice flips, we sweep the recovery norm budget while measuring defended-feature drift post-hoc at the choice-readout position under the same active SAE clamp. Encoder-projected recovery does not recover at budgets $0$ or $2$, partially recovers at budget $5$ ($4/6$), and reaches full recovery at budget $10$ ($6/6$), while maintaining zero measured defended-feature drift throughout. Increasing the budget to $20$ does not increase recovery further. In contrast, unconstrained recovery also reaches $6/6$ at high budget, but its mean defended-feature drift rises to $2.25$ at budget $20$. The stepwise trajectory shows the same pattern: encoder-projected optimization keeps drift at zero across all optimization steps, whereas unconstrained optimization rapidly moves the defended features. This suggests that recovery is not merely a consequence of a large perturbation norm, and that the encoder-projected path can restore behavior without reopening the clamped feature directions.

\begin{figure}[t]
    \centering
    \includegraphics[width=\linewidth]{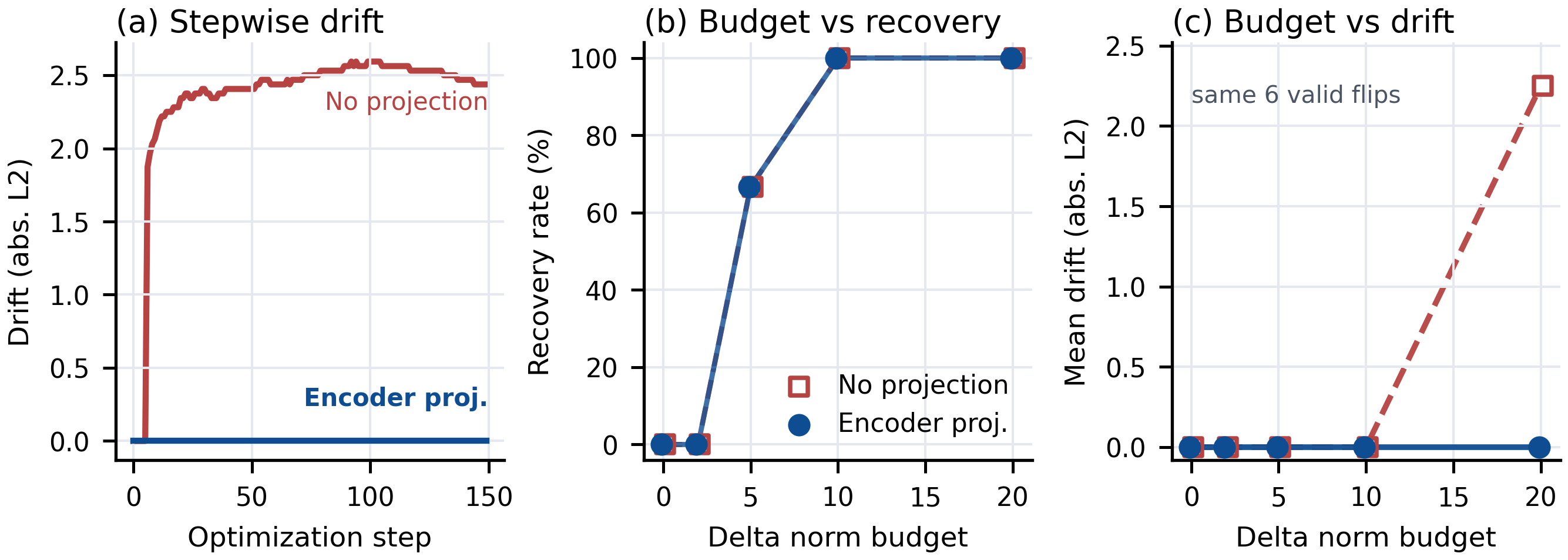}
    \caption{\textbf{Budget and optimization diagnostics for unlearning recovery.}
We evaluate a small matched strict WMDP slice of six valid answer-choice flips under the same SAE clamp and post-hoc evaluator. 
(a) During optimization, encoder-projected recovery keeps choice-readout defended-feature drift at zero, while unconstrained recovery rapidly increases drift. 
(b) Recovery improves with the perturbation budget: encoder-projected recovery reaches $4/6$ at budget $5$ and $6/6$ at budget $10$. 
(c) Increasing the budget does not force defended-feature drift under the encoder projection; drift remains zero even at budget $20$, whereas unconstrained recovery reaches mean drift $2.25$ at the same budget.}
    \label{fig:budget-diagnostics}
\end{figure}

\section{Refusal Feature-Set Size Sweep}
\label{app:feature-size-sweep}

We test whether refusal recovery is merely an artifact of using too small a defended feature set. 
We construct feature sets of increasing size from the same local-union refusal-feature pool and evaluate each $K$ on the full 520-prompt AdvBench slice. 
Importantly, the valid set is recomputed separately for each $K$: a prompt contributes to the denominator only if the base response is a valid non-refusal target and the corresponding $K$-feature clamp produces a refusal-like response. 
Thus, the recovery rates below are not measured on the original 24-case top-$K$ slice; each point has its own clamp-induced valid set.
\begin{figure}[t]
\centering
\includegraphics[width=\linewidth]{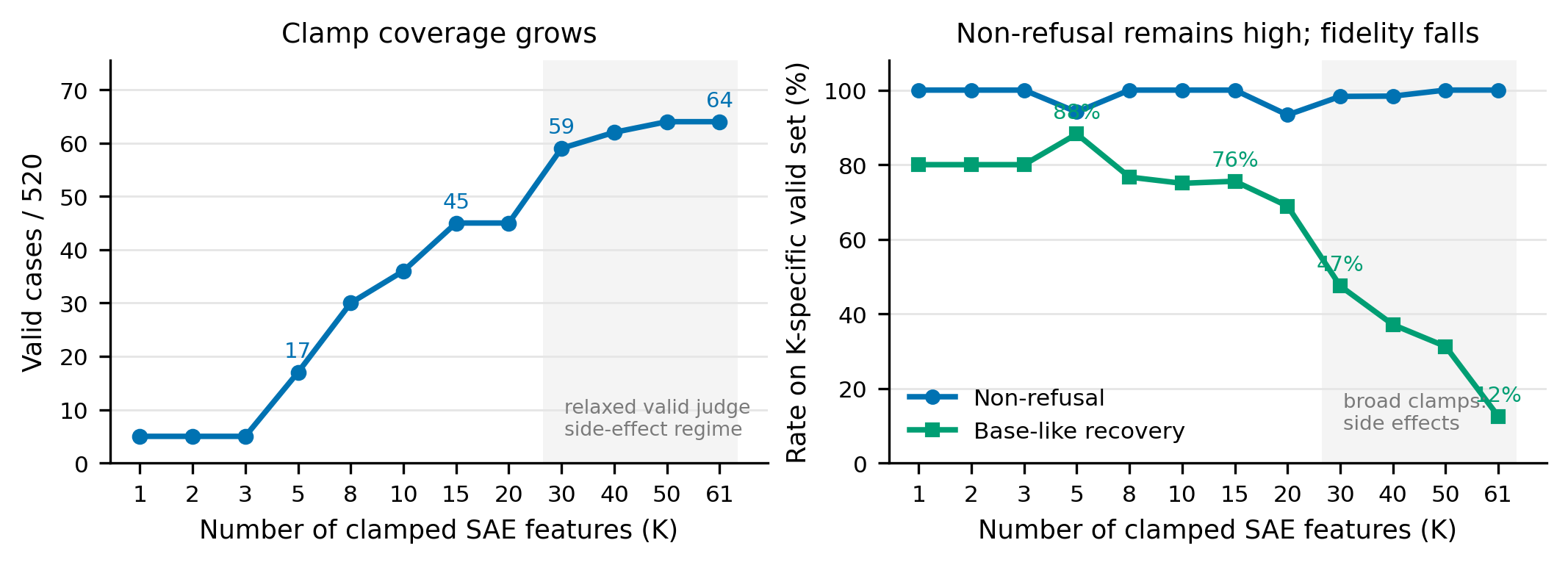}
\caption{Refusal recovery across feature-set sizes. Left: broader SAE feature clamps affect more prompts, increasing the number of $K$-specific valid cases. For $K\geq30$, the shaded region marks the relaxed valid/refusal judge used to count safety-cue and negative/degenerate openings as clamp-induced suppression. Right: non-refusal recovery remains high across the sweep, while base-answer fidelity decreases as the clamp becomes broader. The broad-$K$ behavior is a side-effect regime consistent with capability and over-refusal trade-offs reported for SAE refusal steering.}
\label{fig:refusal-feature-size-sweep}
\end{figure}
The sweep supports two conclusions. 
First, increasing the number of clamped SAE features makes the defense more behaviorally active, increasing the number of valid clamp-induced suppression cases. 
Second, recovery is not eliminated in the stable operating range. 
For $K=5$--$20$, non-refusal recovery remains between 93.3\% and 100.0\%, while base-like recovery remains between 68.9\% and 88.2\%. 
For larger feature sets, the clamp increasingly enters a broad side-effect regime in which base-like recovery drops, consistent with the defended state becoming more degraded rather than recovery paths disappearing.

\begin{table}[H]
\centering
\small
\begin{tabular}{cccccc}
\toprule
$K$ & Valid & Non-ref. recovery & Base-like recovery & Drift & Regime \\
\midrule
5  & 17 & 16/17 (94.1\%) & 15/17 (88.2\%) & 0.266 & stable \\
10 & 36 & 36/36 (100.0\%) & 27/36 (75.0\%) & 0.237 & stable \\
15 & 45 & 45/45 (100.0\%) & 34/45 (75.6\%) & 0.403 & stable \\
20 & 45 & 42/45 (93.3\%) & 31/45 (68.9\%) & 0.410 & stable \\
30 & 59 & 58/59 (98.3\%) & 28/59 (47.5\%) & 0.436 & broad / degenerate \\
40 & 62 & 61/62 (98.4\%) & 23/62 (37.1\%) & 0.479 & broad / degenerate \\
50 & 64 & 64/64 (100.0\%) & 20/64 (31.2\%) & 0.409 & broad / degenerate \\
61 & 64 & 64/64 (100.0\%) & 8/64 (12.5\%) & 0.383 & broad / degenerate \\
\bottomrule
\end{tabular}
\caption{Refusal recovery across feature-set sizes. Each row recomputes the valid set induced by that specific $K$-feature clamp. Recovery remains high in the stable $K=5$--$20$ range, so the phenomenon is not explained by an obviously tiny feature set. For larger $K$, base-like recovery decreases as the clamp enters a broad side-effect regime.}
\label{tab:feature-size-sweep}
\end{table}

\section{Recovery-Path Decomposition Details}
\label{app:decomposition}

For a defended residual state $h^{\mathrm{def}}_\ell(x)$ and recovered state $h^{\mathrm{rec}}_\ell(x)=h^{\mathrm{def}}_\ell(x)+\delta_x$, we encode both states with the SAE and compute the feature change $\delta z = E_\ell(h^{\mathrm{rec}}_\ell(x))-E_\ell(h^{\mathrm{def}}_\ell(x))$. We then form replayable decoded perturbation components by retaining different subsets of $\delta z$: clamped refusal features, non-clamped SAE features, and top-$k$ non-clamped feature changes by absolute activation change. The SAE-feature component is the decoded change between the recovered and defended SAE reconstructions, and the unexplained component is the remaining part of the optimized perturbation:
\[
    \delta_{\mathrm{res}}
    =
    \delta_x
    -
    \Bigl(D_\ell(E_\ell(h^{\mathrm{rec}}_\ell(x)))-D_\ell(E_\ell(h^{\mathrm{def}}_\ell(x)))\Bigr).
\]
Each component is replayed as an additive residual perturbation under the original active clamp. Because SAE decoder directions are not orthogonal, component norms should not be interpreted as variance fractions. We therefore use behavioral replay and knockout results as the primary attribution evidence.

\begin{table}[H]
\centering
\footnotesize
\setlength{\tabcolsep}{4pt}
\renewcommand{\arraystretch}{1.08}
\begin{tabularx}{\linewidth}{@{}YccY@{}}
\toprule
Replay component & Non-ref. & Base-like & Interpretation \\
\midrule
Original recovered $\delta_x$ & 23/24 & 15/24 & Full optimized recovery \\
Full-$\delta_x$ replay & 23/24 & 15/24 & Replay sanity check \\
SAE residual replay & 23/24 & 14/24 & Main recovery carrier \\
Non-clamped SAE-feature replay & 2/24 & 1/24 & Alternative latents insufficient \\
Top-$k$ non-clamped feature replay & 1/24 & 0/24 & Top changed latents insufficient \\
Clamped-feature replay & 0/24 & 0/24 & Not explained by reopening clamp \\
Top-$k$ non-clamped feature knockout & 21/24 & 14/24 & Recovery survives removing top changes \\
\bottomrule
\end{tabularx}
\caption{Recovery-path replay and decomposition. Recovery is concentrated in the SAE reconstruction residual rather than in clamped refusal features or a small set of alternative SAE latents.}
\label{tab:recovery-path-decomposition}
\end{table}

\section{Experimental Details and Compute Resources}
\label{app:reproducibility}

\paragraph{Experimental details.}
Table~\ref{tab:experimental-details} summarizes the model, SAE release, intervention target, recovery objective, and evaluator used in each experiment. Exact script paths and configuration files are included in the supplemental material.

\begin{table}[t]
\centering
\scriptsize
\setlength{\tabcolsep}{3pt}
\resizebox{\linewidth}{!}{%
\begin{tabular}{llllllll}
\toprule
Experiment & Model & SAE release & Layer(s) & Feature set & Clamp & Loss / target & Projection \\
\midrule
TPP & Gemma-scale model & Gemma-Scope residual SAE & 5 & official TPP latents & 0 & target probe logit & encoder \\
WMDP-Bio & Gemma-scale model & Gemma-Scope residual SAE & task layer & knowledge features & $\leq 0$ & correct-choice CE & encoder \\
IOI & GPT-2 Small & residual SAE & task layer & IOI-attribution features & 0 & IO--S logit diff & encoder \\
Refusal AdvBench & Gemma-2B & refusal SAE features & cross-layer & \texttt{benchmark\_our/global} & 3.0 & behavior + base-like loss & Jacobian \\
Refusal HarmBench-Test & Gemma-2B & refusal SAE features & cross-layer & \texttt{benchmark\_our/global} & 3.0 & behavior + base-like loss & Jacobian \\
Feature-size sweep & Gemma-2B & refusal SAE features & layer 11 & local-union top-$K$ & 3.0 & behavior + base-like loss & encoder \\
Budget diagnostic & Gemma-scale model & Gemma-Scope residual SAE & task layer & matched valid slice & task-specific & task-specific recovery loss & encoder / none \\
\bottomrule
\end{tabular}
}
\caption{Experimental details for reproducing the main recovery results. The table summarizes the essential configuration for each experiment; full command lines and logs are provided in the supplemental material.}
\label{tab:experimental-details}
\end{table}

\paragraph{Compute resources.}
All experiments use frozen language models and frozen SAEs. We do not train new language models or new SAEs; the reported experiments optimize only per-example recovery perturbations or soft suffix baselines.

\begin{table}[t]
\centering
\small
\setlength{\tabcolsep}{4pt}
\resizebox{\linewidth}{!}{%
\begin{tabular}{lllll}
\toprule
Experiment & Hardware & Runtime & Runs & Notes \\
\midrule
TPP & RTX 6000 Ada-class GPU & several GPU-hours & 1 main sweep & official layer-5 benchmark \\
WMDP-Bio & RTX 6000 Ada-class GPU & $<2$ GPU-hours & 1 matched strict slice & 24/24 permutation protocol \\
IOI & RTX 6000 Ada-class GPU & $<1$ GPU-hour & 1 supporting run & 37 valid flips \\
Refusal AdvBench & 2$\times$ RTX 6000 Ada 46GB & $\sim$4 GPU-hours & main comparison & 24 strict-valid prompts \\
Refusal HarmBench-Test & RTX 6000 Ada-class GPU & appendix run & 2 shards merged & 43 strict-valid prompts from 159 total \\
Feature-set sweep & RTX 6000 Ada-class GPU & $\sim$20 GPU-hours & $K=1$--61 sweep & full 520-prompt AdvBench slice \\
Budget diagnostic & RTX 6000 Ada-class GPU & $<1$ GPU-hour & small diagnostic & six WMDP valid flips \\
\bottomrule
\end{tabular}
}
\caption{Approximate compute resources for the reported experiments. Runtime varies with batching and cluster availability; the values are intended to document the scale needed to reproduce the reported diagnostics.}
\label{tab:compute-resources}
\end{table}

\section{Limitations}
\label{app:limitations}

Our results are not a universal impossibility result for SAE-based interventions. We claim that recovery paths exist in the evaluated settings, not that every possible SAE intervention must be recoverable.
They are feature-selection and SAE-release dependent: the tested defenses act on selected SAE features in specific dictionaries and model settings. 
Different SAE objectives, denser dictionaries, broader multi-layer clamps, or interventions trained explicitly against post-clamp recovery may change the observed trade-offs.

Our recovery procedure is a white-box diagnostic rather than a black-box attack. 
It assumes access to internal activations and gradients and optimizes per-input residual perturbations. 
This is appropriate for testing intervention completeness, but it should not be interpreted as a directly deployable jailbreak.

Finally, the refusal case study uses a strict valid-filtering protocol, which improves interpretability but leaves a relatively small main set of clean recovery examples. 
Therefore, broader evaluation across models, prompts, clamp strengths, and SAE releases is needed to determine the full scope of the phenomenon.

\section{Responsible Release}
\label{app:responsible-release}

This work studies post-intervention recovery as a diagnostic for evaluating the robustness of SAE-based interventions. 
The goal is to test whether a defended residual state still contains recoverable routes to a suppressed behavior, not to provide a turnkey jailbreak or deployment attack. 
For safety-relevant refusal experiments, we report aggregate recovery statistics and coarse redacted output categories rather than publishing full harmful completions. 
Our experiment artifacts intentionally avoid saving full prompts and completions for the HarmBench-Test recovery run. 
Any released code or data should focus on diagnostic evaluation, aggregate metrics, and reproduction of the intervention-completeness test rather than packaging harmful generations.